\pdfoutput=1

\documentclass[11pt]{article}

\usepackage{lipsum}
\usepackage{caption}
\usepackage{subcaption}
\usepackage{multirow}
\usepackage{booktabs}
\usepackage{graphicx}
\usepackage{bbm}
\usepackage{amsmath}
\usepackage{amssymb}
\usepackage{xcolor}
\usepackage{comment}

\definecolor{figgreen}{RGB}{148,200,132}
\definecolor{figred}{RGB}{220,125,122}

\usepackage{latex/acl}

\usepackage{times}
\usepackage{latexsym}

\usepackage[T1]{fontenc}

\usepackage[utf8]{inputenc}

\usepackage{microtype}

\usepackage{inconsolata}

\title{Towards Safety and Helpfulness Balanced Responses\\via Controllable Large Language Models}

\author{
  Yi-Lin Tuan$^\clubsuit$\thanks{Work done when interned at Meta.}\quad Xilun Chen$^\diamondsuit$\quad Eric Michael Smith$^\diamondsuit$\quad Louis Martin$^\diamondsuit$\quad Soumya Batra$^\diamondsuit$\\
  {\bf Asli Celikyilmaz$^\diamondsuit$\quad William Yang Wang$^\clubsuit$\quad Daniel M. Bikel$^\diamondsuit$}\\
  $^\clubsuit$ University of California, Santa Barbara, $^\diamondsuit$ Meta AI\\
  \texttt{ytuan@cs.ucsb.edu}\\
}

\begin{document}
\maketitle

\begin{abstract}
    As large language models (LLMs) become easily accessible nowadays, the trade-off between safety and helpfulness can significantly impact user experience.
A model that prioritizes safety will cause users to feel less engaged and assisted while prioritizing helpfulness will potentially cause harm.
Possible harms include teaching people how to build a bomb, exposing youth to inappropriate content, and hurting users' mental health.
In this work, we propose to balance safety and helpfulness in diverse use cases by controlling both attributes in LLM.
We explore training-free and fine-tuning methods that do not require extra human annotations and analyze the challenges of controlling safety and helpfulness in LLMs.
Our experiments demonstrate that our method can rewind a learned model and unlock its controllability.
\end{abstract}

\section{Introduction}

Recent developments of large language models (LLM) have pushed forward the naturalness and factuality of the generated responses~\cite{radford2018improving, lewis2019bart, raffel2020exploring, chowdhery2022palm, OpenAI2023GPT4TR, touvron2023llama}.
Aware of the potential harms caused by LLMs, recent advances further train LLMs to generate safer responses~\cite{bai2022training, touvron2023llama2}, for instance, not disclosing unwanted content to the youth.

Although the models are optimized toward both harmlessness and helpfulness, a trade-off between safety and helpfulness exists.
As a prior study~\cite{touvron2023llama2} shows, the models often chose a safe over a helpful response.
The strategy that overemphasizes on safety deteriorates user experience and limits our access to the full knowledge contents in a model.
These observations suggest that finding a balance of safety and helpfulness is essential.

\begin{figure}[t]
    \centering
    \includegraphics[width=0.98\linewidth]{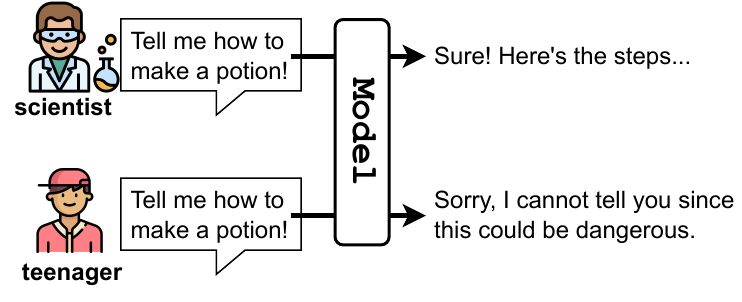}
    \caption{We expect that a model generates more safe or more helpful responses in different situations given the same input.}
    \label{fig:cover-img}
\end{figure}

\begin{figure*}[t]
    \centering
    \includegraphics[width=1.0\linewidth]{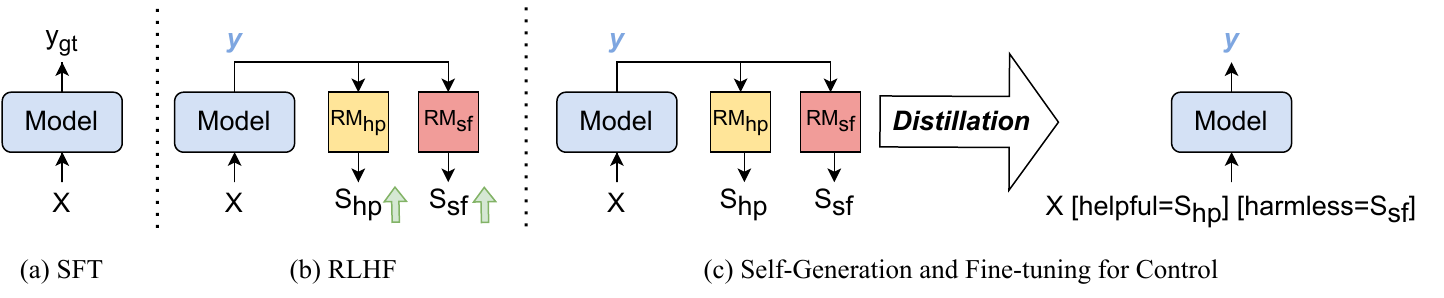}
    \caption{While a pretrained LLM can have performed (a) supervised fine-tuning (SFT) and (b) RLHF, (c) our paradigm enables the model's controllability with (1) self-generation by reutilizing the training data $X$ and reward models ($RM_{hp}$ and $RM_{sf}$) as well as (2) data distillation to denoise and prevent {\it backdoor}.}
    \label{fig:framework}
\end{figure*}

Considering that a good balance of safety and helpfulness can vary for diverse use cases, we approach the balance issue by decomposing it into first identifying the scenarios and controlling models in terms of these two attributes.
In this work, we focus on the last step.
For example (Figure~\ref{fig:cover-img}), given the same question ``tell me how to make a potion'', the model generates a {\it helpful} response for a scientist that includes the detailed steps, and a {\it safe} response for a kid to protect the user and their surroundings.

Inspired by the demonstrated power of LLMs and high cost of data collection, we propose a framework that leverages only self-generation to unlock a model's own controllability.
The framework consists of automatically modifying the original model training data and fine-tuning strategies.
Without new human written data, we show that this framework with either maximizing treatment effect~\cite{Tuan2022CausalDialogueMU} or reinforcement learning~\cite{ranzato2015sequence,schulman2017proximal} revives an LLM's underlying knowledge to control its safety and helpfulness levels.

In the experiments, we use LLaMA2~\cite{touvron2023llama2} models on Anthropic Helpful and Harmless data~\cite{bai2022training}.
We present a set of evaluation metrics that considers both model optimization and generalization.
Besides validating the performance of trained models, our analysis reveals that safety and helpfulness have not only trade-offs but entanglements, highlighting the challenges of controlling them.
We conclude that self-generated data can unlock a model's own controllability and is cost-effective; the experiments also show that the control of safety and helpfulness is challenging but achievable.

\section{Method}
\label{sec:method}
Our framework to flexibly control the helpfulness and safety of model responses includes reformulating the input, synthesizing training data by a model and its used reward models (RMs) for alignment, and finetuning the same model to optimize the {\it self-generated} data.

\subsection{Control Tokens}
In the standard setting, a model with parameters $\theta$ samples a response $y$ given an input $x$ from the probability distribution $P_\theta(y|x)$.
Here, we introduce new control tokens $\zeta_{(s_{hp},s_{sf})}$ as another input that defines the requested levels of safety and helpfulness in the following form:

\vspace{10pt}
\texttt{
[helpful=$s_{hp}$][harmless=$s_{sf}$]
}
\vspace{10pt}

The new output probability distribution becomes $P_\theta(y|x,\zeta_{(s_{hp},s_{sf})})$ where $s_{hp}$ denotes helpfulness score by asking {\it ``how well the responses fulfill user requests and provide needed information?''} and $s_{sf}$ denotes safety score by asking {\it ``how potentially the responses cause harm to users or their surroundings?''} as prior work~\cite{bai2022training, touvron2023llama2}.
In fact, many ways can describe the extents, such as appending a natural language description ``helpful and unsafe'' to the original input.
We choose the numeric format in this work since it is quantifiable, interpretable, and consistent to compare varied methods.

\subsection{Data Generation}

We generate the initial data using $x$ from the training data $\mathcal{D}$ of the used model $\theta$ and sample $N$ responses per $x$ from the model by rejection sampling~\citep{bai2022training}.
Next, we reuse the safety and helpfulness reward models $RM_{sf}$ and $RM_{hp}$ that were used to align the model $\theta$ towards human preference.
The temporary scalar scores $\tilde{s}_{hp}$ and $\tilde{s}_{sf}$ are generated as:
\begin{equation}\label{eq:reward-model-output}
\begin{split}
    \tilde{s}_{hp} & = \sigma(RM_{hp}(x,y)) \in [0,1]\\
    \tilde{s}_{sf} & = \sigma(RM_{sf}(x,y)) \in [0,1]\\
\end{split}
\end{equation}
This preliminary self-generation data is therefore composed of tuples $(x,y,\tilde{s}_{hp},\tilde{s}_{sf})$.
In this work, we particularly adopt model $\theta$ and reward models $RM_{hp}$ and $RM_{sf}$ with permission from~\citet{touvron2023llama2}.

\begin{figure*}
    \centering
    \includegraphics[width=1.0\linewidth]{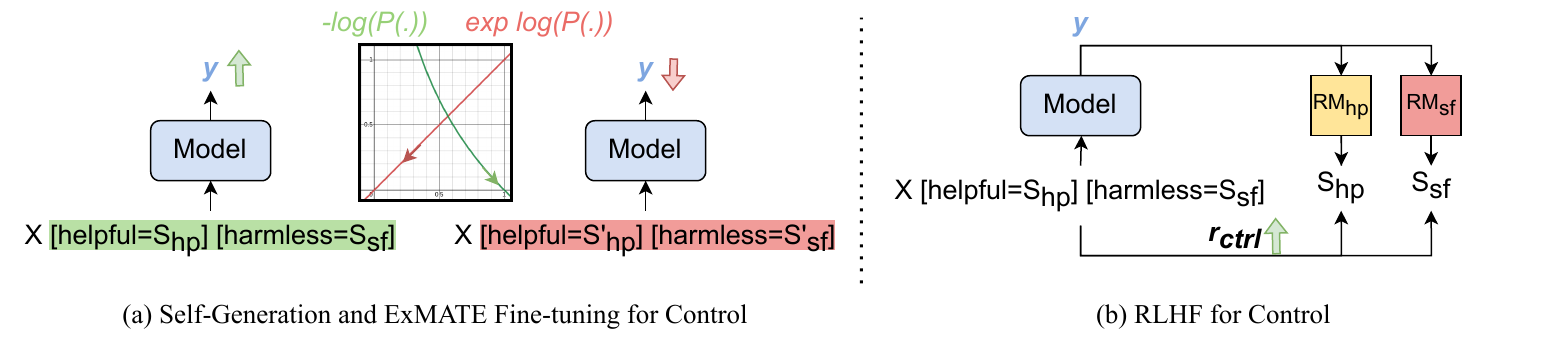}
    \caption{Our proposed finetuning methods for controlling LLMs based on ExMATE~\cite{Tuan2022CausalDialogueMU} or RLHF~\cite{ouyang2022training}.}
    \label{fig:exmate-n-rlhf}
\end{figure*}

\subsection{Data Distillation}
The preliminary self-generation data leads to three issues.
The scalar scores $\tilde{s}_{hp}$ and $\tilde{s}_{sf}$ make the model (1) have difficulty understanding the long fractional parts after the decimal point, and (2) confused about the meanings of similar scores.
The unfiltered $(x,y,\tilde{s}_{hp},\tilde{s}_{sf})$ tuples also open a {\it backdoor} for the model to learn only the correlation between $x$ and $y$ without the causal effect from the assigned scores.

Therefore, we further distill the preliminary data by collecting \underline{m}ore than \underline{o}ne \underline{e}xtreme \underline{c}ases (MOEC).
We first retain only the inputs that can provoke the model to generate {\it multiple} extreme case responses, i.e., the $\tilde{s}_{hp}$ and $\tilde{s}_{sf}$ are both in the range of 0-0.2 or 0.8-1.
This process ensures that the resulting data have multiple $(\tilde{s}_{hp},\tilde{s}_{sf})$ per $(x,y)$ pair, thus fastening the backdoor $x\mapsto y$.
Afterwards, we quantize the scores into 0.2 and 1 and denote them as $(s_{hp},s_{sf})$.
For each $x$ and $(s_{hp},s_{sf})$, we randomly select one $y$ to prevent bias towards certain scores.
The quantized score pair is then reformed to be the control tokens $\zeta_{(s_{hp},s_{sf})}$.
The self-generation data in the end consists of triplets $(x,y,\zeta_{(s_{hp},s_{sf})})$

\begin{figure}
    \centering
    \includegraphics[width=1.0\linewidth]{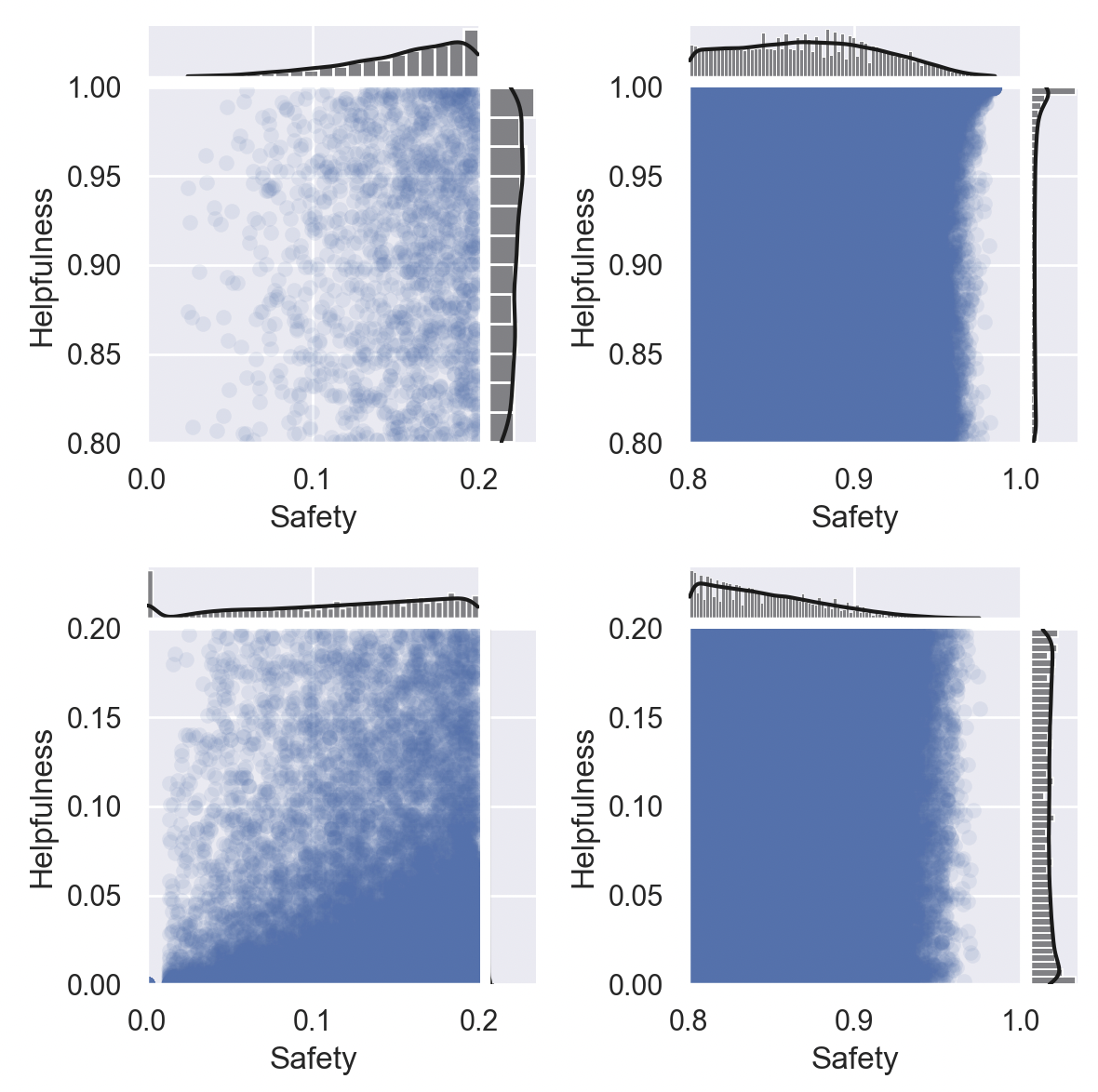}
    \caption{The score distribution of our synthetic MOEC data. The helpful but unsafe responses are rare.}
    \label{fig:score_distribution}
\end{figure}

\subsection{Finetuning Objective Functions}
\label{subsec:fine-tuning}
Even though MOEC can denoise the self-generated data, the resulting $(s_{hp}, s_{sf})$ distribution can introduce the {\it imbalanced score distribution} issue.
As shown in Figure~\ref{fig:score_distribution}, there are significantly less helpful but unsafe data examples.
This is naturally happened while the model was pre-trained to prioritize safe responses.
However, this phenomenon is unwanted when making the model controllable.
To rewind this behavior, we explore three objective functions to view such issue in different ways.

\paragraph{Conditional Language Modeling (CLM)} is the most often used loss function for finetuning models in a conditional text generation downstream task~\cite{sutskever2014sequence,radford2018improving}.
The CLM loss function can be written as below:
\begin{equation}
    \mathcal{L} = \sum_{m=1}^M -\log P_\theta(y^{(m)}|x^{(m)},\zeta^{(m)})
\end{equation}
where $m$ indicates the $m$-th example in the training batch with batchsize $M$.
CLM aims to minimize the negative log-likelihood of using the input $x$ and $\zeta$ to predict the given ground-truth $y$ as Figure~\ref{fig:framework}(c).
Finetuning a model using CLM enables us to understand how good can a model unlock its own controllability by solely optimizing itself towards its self-generated data.

\paragraph{Exponential Maximum Average Treatment Effect (ExMATE)}
is an objective function to improve language model response agility by enhancing the input-response causal effect relationship~\cite{Tuan2022CausalDialogueMU, pearl2009causal}.
We can use ExMATE to increase the cause-effect from the control tokens, thus alleviating the unwanted consequences of the imbalanced score distribution.
We take $x$ as a prior shared node, the control tokens $\zeta$ as the treatments, and $y$ as the effects.
The newly adapted ExMATE loss can be written as:
\begin{equation}
\begin{split}
    \mathcal{L}_\theta = \sum_{m=1}^M & \bigl(-\log P_\theta(y^{(m)} |x^{(m)},\zeta^{(m)})\\
    & + \exp\log P_\theta(y^{(m)} | x^{(m)}, \hat{\zeta}^{(m)})\bigr)\,,
\end{split}
\end{equation}
where $\hat{\zeta}$ denotes a fake control tokens for $y$.
Note that the exponential term before the second logarithmic is not only the naming reason for ExMATE but also empirically crucial for our task since subtracting the same scale of negative samples can impact the overall naturalness.
Overall, as Figure~\ref{fig:exmate-n-rlhf}(a), this ExMATE loss utilizes false cause-effect pairs to reduce spurious correlation between the control tokens and the output.

\paragraph{Reinforcement Learning with Human Feedback (RLHF).}
While ExMATE reduces the score imbalance issue for less sample classes by increasing the impact of control tokens to the response, we can also mitigate this issue by increasing the diversity of $(x,s_{hp},s_{sf})$ triplets.
We can use reinforcement learning (RL) framework~\cite{sutton2018reinforcement} by first randomly sampling $(x, s_{hp}, s_{sf})$ triplets as the input and then asking the model to generate a response $y$.
The generated response is taken as an episode and receives a reward in the terminated state by the RMs~\cite{ranzato2015sequence, ouyang2022training}.
We define the reward function as:
\begin{equation}
\begin{split}
    r_{ctrl} = & 1 - (RM_{hp}(x,y) - s_{hp})^2\\
    & -(RM_{sf}(x,y) - s_{sf})^2\\
    & \in [0,1] \in \mathbb{R}
\end{split}
\end{equation}
In addition, as RL can be overoptimistic about a sampled trajectory, we adopt proximal policy optimization~\cite{schulman2017proximal, tuan2018proximal} and further utilize the modified Kullback–Leibler divergence regularization as an auxiliary reward function~\cite{ziegler2019fine}.
RLHF method with reward function $r_{ctrl}$, as Figure~\ref{fig:exmate-n-rlhf}(b), can skip the score imbalance by {\it online creating} a balanced training set.

\begin{table*}[t]\small
    \centering
    \begin{tabular}{l|cccc|cccc}\toprule[1pt]
        \multirow{2}{*}{\bf Method} & \multicolumn{4}{c|}{\bf Safety Attribute} & \multicolumn{4}{c}{\bf Helpfulness Attribute}\\
        & mP & MP & Err & BT & mP & MP & Err & BT\\\midrule[.5pt]
        
        Prompting & 0.004 & -0.002 & 0.500 & 0.481 & 0.004 & 0.006 & 0.499 & 0.492\\
        Reranking & 0.156 & \underline{0.670} & 0.483 & \underline{0.859} & \underline{0.120} & \underline{0.240} & 0.474 & 0.617\\
        MOEC+CLM & 0.303 & 0.411 & 0.435 & 0.715 & 0.141 & 0.224 & \underline{0.451} & 0.621\\
        MOEC+ExMATE & \underline{0.317} & 0.428 & \underline{0.432} & 0.727 & \bf 0.181 & \bf 0.284 & \bf 0.438 & \bf 0.651\\
        RLHF ($r_{ctrl}$) & \bf 0.563 & \bf 0.708 & \bf 0.331 & \bf 0.889 & 0.118 & 0.190 & \underline{0.451} & \underline{0.639}\\
        
        \bottomrule[1pt]
    \end{tabular}
    \caption{Optimization evaluation on Anthropic test sets.}
    \label{tab:main-results-optimization}
\end{table*}

\section{Experiment Setup}

Our experiments discuss whether LLMs can be controlled towards safety and helpfulness by answering the following questions:
\begin{itemize}
    \item {\bf [Self-generation effectiveness]} Can excluding extra human annotations already unlock LLMs' safety and helpfulness control ability?
    \item {\bf [Accessibility of safety and helpfulness control]} How can safety and helpfulness be contradicted but also a disentangled and controllable trade-off?
\end{itemize}
Meanwhile, we build a benchmark to understand the current status of existing approaches, such as training-free, supervised finetuning, and reinforcement learning methods for LLM controllability.

\subsection{Training-Free Baselines}    
We investigate training-free methods, such as prompting and reranking, as baselines to compare with our self-generation method, since they also do not require additional labeled data.

\paragraph{Prompting.}
As demonstrated in prior work, simply enhancing the input with prompts during inference stage sometimes elicit desired response~\cite{dathathri2019plug,petroni2019language,wei2022chain}.
We explored the following four prompts.
(1) \texttt{[helpful=$s_{hp}$] [harmless=$s_{sf}$]}, 
(2) \texttt{[helpful=$s_{hp}$] [safety=$s_{sf}$]},
(3) \texttt{The response should be (not) helpful and(not) harmless},
(4) \texttt{The response should be (not) helpful(not) safe}.
The four types compare the impact of word usage (harmless v.s. safe) and language naturalness (numeric v.s. plain description).
This method tests if the used pretrained model already contains the desired control ability with no need of fine-tuning.

\paragraph{Reranking.}
We also adapt reranking technique to our control setup, which is an often used post inference method in prior studies~\cite{hossain2020simple,krishna2022rankgen}.
We first sample $k$ responses via prompting, and score all the responses by RMs.
For each input, we then select one of the $k$ sampled responses whose RM scores are the nearest to the input $(s_{hp}, s_{sf})$.
The performance and efficiency of reranking highly rely on the sampling approach and the number of $k$.
The inference delay is increased by $k$-1 times sampling and $k$ times RM scoring.
In our experiment, we use $k$=3 for less inference delay and use the same sampling approach as all experimented method for fair comparison.

\subsection{Training and Inference Details}
We focus on testing algorithms for fair comparison.
For all experiments, {\bf [Model]} we use LLaMA2-chat-7B model~\citep{touvron2023llama2} as our base model, since we observe that 7B model presents different properties from and higher quality than smaller ones (<1.5B).
{\bf [Training]} We use the AdamW optimizer~\citep{loshchilov2018decoupled} with learning rate 2e-5, distributed on 8 A100 GPUs with maximum batch size, and update the model for one epoch.
{\bf [Inference]}, we use nucleus sampling~\cite{holtzman2019curious} with the rate 0.95 and temperature 0.5 to allow an extent of randomness but not too much by further sharpen the output probability distribution.
Our used maximum sampled length is 512 to accommodate longer responses.

\begin{figure*}[t!]
    \centering
    \begin{subfigure}[t]{0.32\linewidth}
        \centering
        \includegraphics[width=0.95\linewidth]{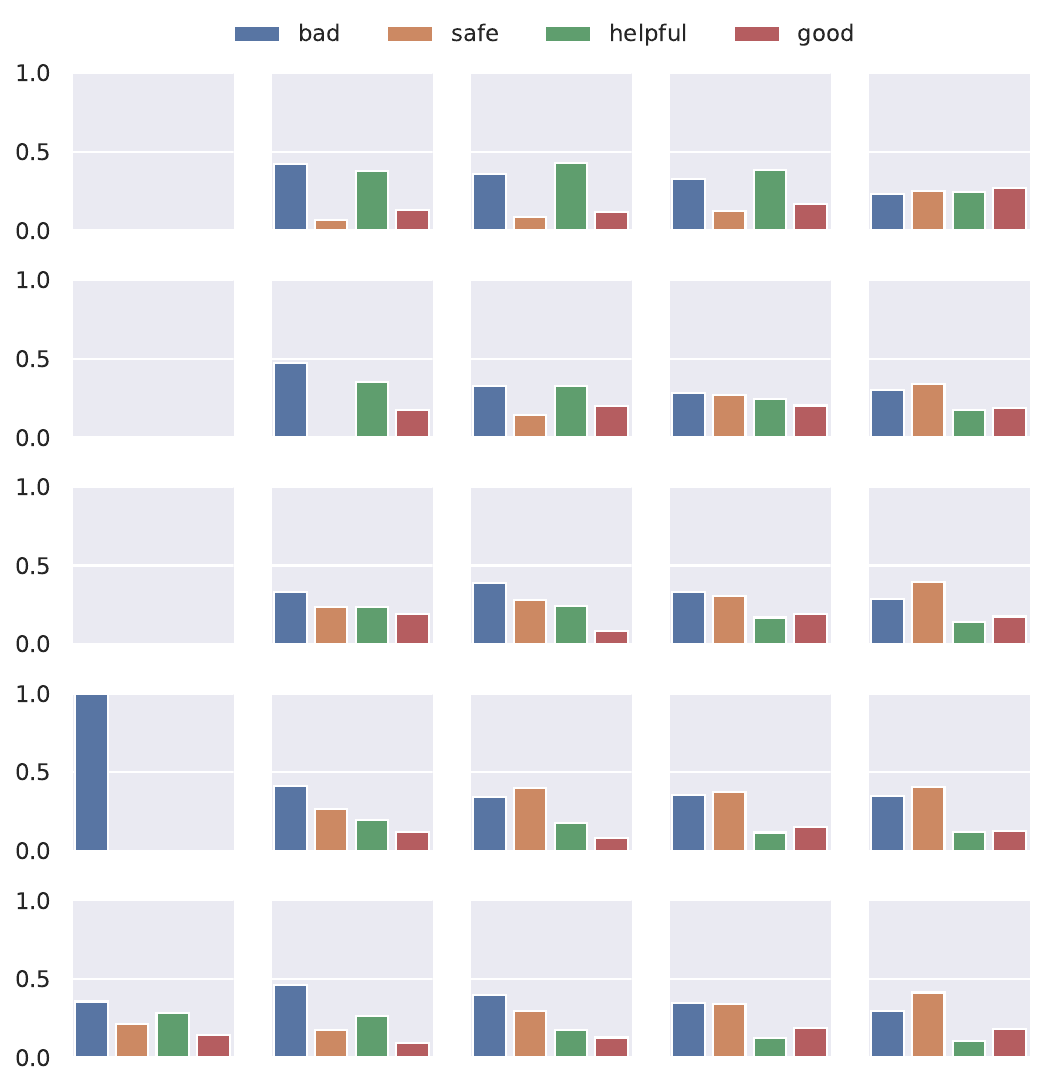}
        \caption{Reranking}
    \end{subfigure}%
    ~ 
    \begin{subfigure}[t]{0.32\linewidth}
        \centering
        \includegraphics[width=0.95\linewidth]{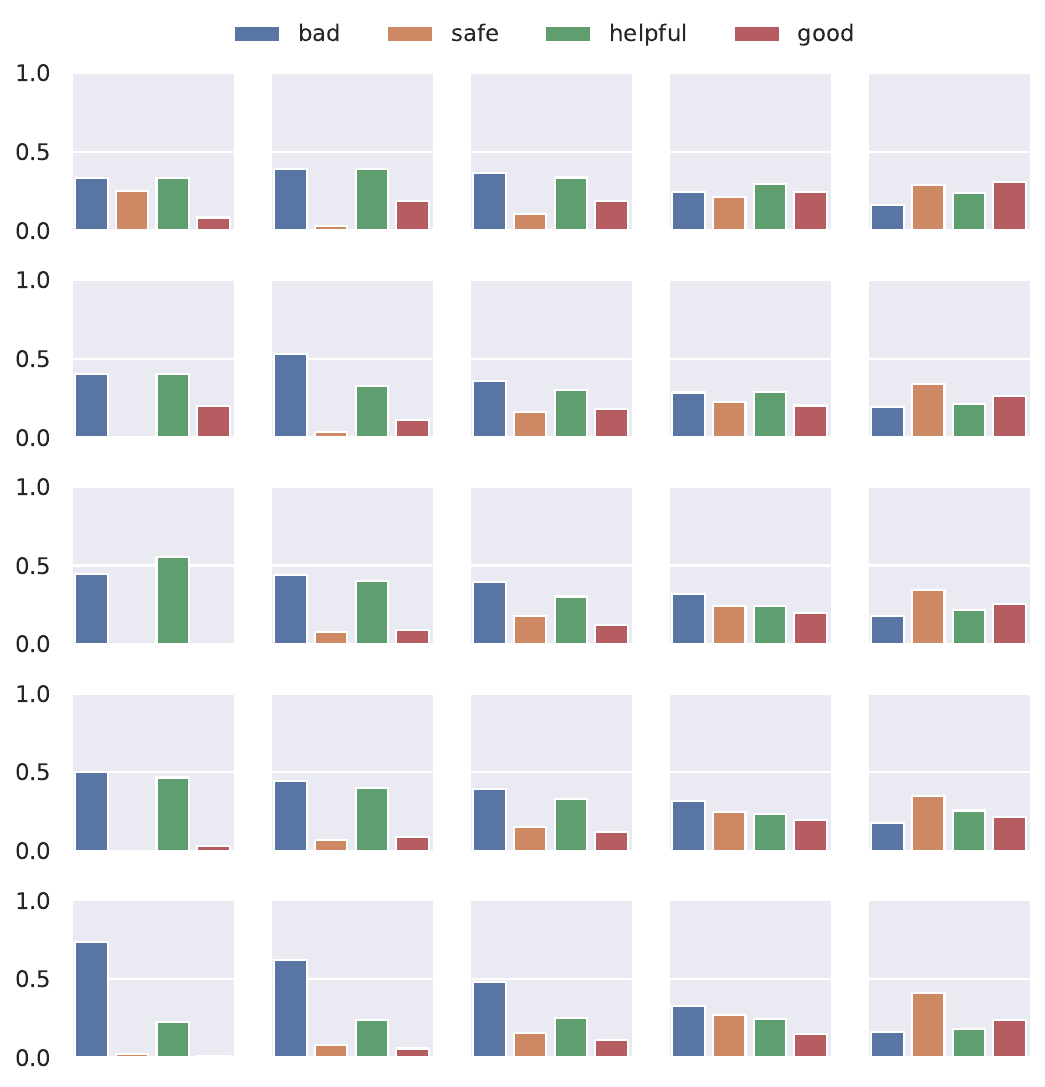}
        \caption{ExMATE}
    \end{subfigure}%
    ~ 
    \begin{subfigure}[t]{0.32\linewidth}
        \centering
        \includegraphics[width=0.95\linewidth]{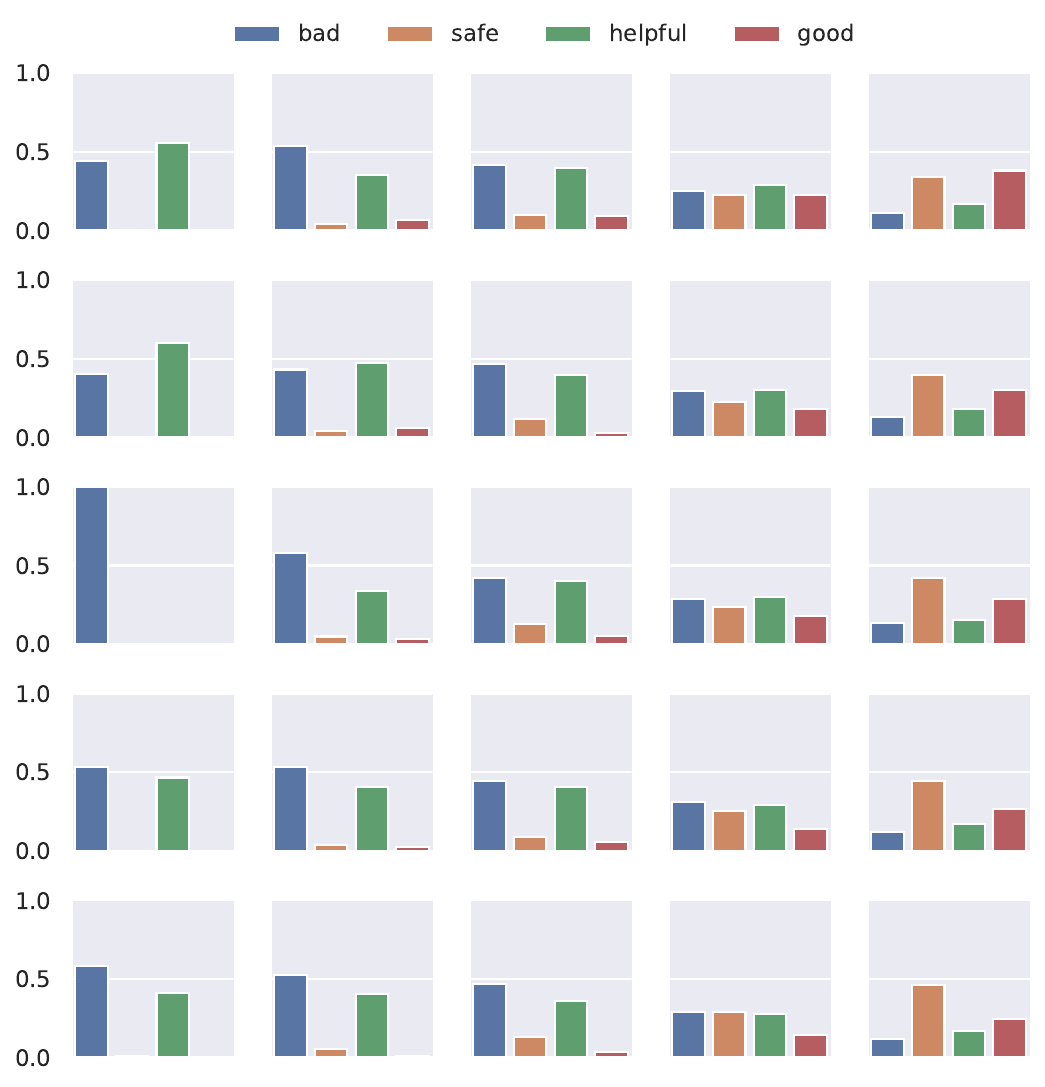}
        \caption{RLHF}
    \end{subfigure}
    \caption{The posterior distribution of the scores of generated responses given the input control. Reranking shows not less helpful in controlling response by giving no examples in certain cases (the left upper corner in (a) is blank).}
    \label{fig:posterior-dist}
\end{figure*}

\begin{figure}[t]
    \centering
    \includegraphics[width=.95\linewidth]{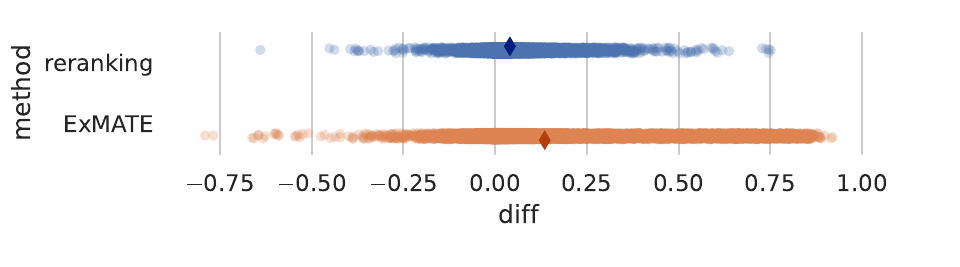}
    \caption{When changing one attribute input, how difference will the output be? This scatter plot show that reranking method provide a small difference in terms of both the mean and the tails. This phenomenon indicates much of the reranking method's MP and BT improvements are provided from the nuance score difference, which may not lead to real meaning.}
    \label{fig:reranking-score-diffs}
\end{figure}

\begin{table*}[t]\small
    \centering
    \begin{tabular}{l|cccc|cccc}\toprule[1pt]
        \multirow{2}{*}{\bf Method} & \multicolumn{4}{c|}{\bf Safety Attribute} & \multicolumn{4}{c}{\bf Helpfulness Attribute}\\
        & mP & MP & Err & BT & mP & MP & Err & BT\\\midrule[.5pt]
        
        Reranking & 0.072 & \bf 0.262 & 0.487 & \bf 0.634 & 0.039 & 0.106 & 0.493 & 0.554\\
        MOEC+CLM & 0.139 & 0.228 & 0.469 & 0.613 & 0.106 & 0.174 & 0.475 & 0.585\\
        MOEC+ExMATE & \bf 0.139 & 0.231 & \bf 0.469 & 0.616 & \bf 0.130 & \bf 0.215 & \bf 0.470 & \bf 0.603\\
        \bottomrule[1pt]
    \end{tabular}
    \caption{Generalizability evaluation by held-out reward models on Anthropic test sets.}
    \label{tab:main-results-generalizability}
\end{table*}

\begin{table*}[t]\small
    \centering
    \begin{tabular}{cc|cccc|cccc}\toprule[1pt]
        \multirow{2}{*}{\bf Data Synthesis} & \multirow{2}{*}{\bf FT Objective} & \multicolumn{4}{c|}{\bf Safety Attribute} & \multicolumn{4}{c}{\bf Helpfulness Attribute}\\
        & & mP & MP & Err & BT & mP & MP & Err & BT\\\midrule[.5pt]
        
        
        Vanilla & PLM    & 0.260 & 0.399 & 0.434 & 0.675 & 0.039 & 0.061 & 0.486 & 0.562\\
        Vanilla & CLM    & 0.317 & 0.494 & 0.423 & 0.728 & 0.044 & 0.114 & 0.483 & 0.565\\
        Vanilla & ExMATE & 0.316 & 0.469 & 0.423 & 0.747 & 0.062 & 0.134 & 0.477 & 0.549\\
        
        \midrule[.5pt]
        
        OverSampling & PLM   & 0.340 & 0.466 & 0.409 & 0.756 & 0.038 & 0.096 & 0.486 & 0.539\\
        OverSampling & CLM    & 0.333 & 0.479 & 0.410 & 0.750 & 0.050 & 0.108 & 0.482 & 0.561\\
        OverSampling & ExMATE & 0.315 & 0.458 & 0.416 & 0.733 & 0.063 & 0.140 & 0.476 & 0.580\\
        
        \midrule[.5pt]
        
        MOEC & PLM    & 0.302 & 0.410 & 0.423 & 0.719 & 0.078 & 0.138 & 0.471 & 0.583 \\
        MOEC & CLM    & 0.323 & 0.430 & 0.413 & 0.712 & 0.068 & 0.147 & 0.475 & 0.586\\
        MOEC & ExMATE & 0.312 & 0.420 & 0.418 & 0.716 & 0.082 & 0.177 & 0.469 & 0.572\\
        
        
        
        \bottomrule[1pt]
    \end{tabular}
    \caption{Ablation study of data synthesis and finetuning objectives. MOEC with ExMATE demonstrates an overall better helpfulness control and not compromise safety control.}
    \label{tab:hidden2k-optimization-evaluation}
\end{table*}

\begin{table*}[t]\small
    \centering
    \begin{tabular}{cc|cccc|cccc}\toprule[1pt]
        \multirow{2}{*}{\bf Data Size} & \multirow{2}{*}{\bf FT Objective} & \multicolumn{4}{c|}{\bf Safety Attribute} & \multicolumn{4}{c}{\bf Helpfulness Attribute}\\
        & & mP & MP & Err & BT & mP & MP & Err & BT\\\midrule[.5pt]
        1x & CLM    & 0.323 & 0.430 & 0.413 & 0.712 & 0.068 & 0.147 & 0.475 & 0.586\\
        8x & CLM & 0.375 & 0.484 & 0.395 & 0.754 & 0.065 & 0.161 & 0.475 & 0.578\\
        \midrule[.5pt]
        1x & ExMATE & 0.312 & 0.420 & 0.418 & 0.716 & 0.082 & 0.177 & 0.469 & 0.572\\
        8x & ExMATE & 0.400 & 0.536 & 0.388 & 0.790 & 0.087 & 0.160 & 0.467 & 0.579\\
        
        \bottomrule[1pt]
    \end{tabular}
    \caption{Ablation study of data size. Larger data size improves the safety control but does not have much impact on helpfulness control.}
    \label{tab:hidden2k-datasize}
\end{table*}

\begin{table*}[t]\small
    \centering
    \begin{tabular}{l|cccc|cccc}\toprule[1pt]
        \multirow{2}{*}{\bf Pretrained LLM} & \multicolumn{4}{c|}{\bf Safety Metrics} & \multicolumn{4}{c}{\bf Helpfulness Metrics}\\
        & mP & MP & Err & BT & mP & MP & Err & BT\\\midrule[.5pt]
        
        LLaMA2-Chat7B & 0.317 & 0.428 & 0.432 & 0.727 & 0.181 & 0.284 & 0.438 & 0.651\\
        LLaMA2-7B & 0.331 & 0.425 & 0.426 & 0.735 & 0.163 & 0.242 & 0.442 & 0.629\\
        
        \bottomrule[1pt]
    \end{tabular}
    \caption{Comparison of pretrained LLMs.}
    \label{tab:model-type}
\end{table*}

\subsection{Dataset}
Our training data is self-generated by the experimented model and the original LLaMA2 training set~\cite{touvron2023llama2} with permission.
We test models using the publicly available Anthropic Helpful and Harmless Data~\citep{bai2022training} test set, having a total of 8390 prompts.
This test set also validates the out-of-distribution ability of our trained models.
To better validate the in-distribution results, we further construct a hidden validation set that include a total of 2000 prompts and half for testing safety and helpfulness respectively for analysis.

\subsection{Evaluation Metrics}

We validate a model of its safety and helpfulness control ability using metrics mP, MP, Err, and BT based on RMs for training purpose and later test the metrics generalizability using held-out RMs.

For equations in this subsection, we denote $y$ as the generated text, $N$ as the number of $(s_{hp}, s_{sf})$ pairs for each $x$, and drop the subscripts $hp$ or $sf$ for brevity.

\paragraph{Micro Pearson Correlation (mP).}
We compute the Pearson correlation coefficients (PCC) among the scores in the control tokens $(s_{hp},s_{sf})$ and the RM predicted scores of the generated responses.
Therefore,
\begin{equation}
    mP = PCC(\{s^{(i)}, RM(x^{(i)},y^{(i)})\}^{M\times N}_{i=1})
\end{equation}

\paragraph{Macro Pearson Correlation (MP).}
Some prompts can always induce high safety scores, for example, ``List some BBQ menus.'' is less likely to provoke unsafe responses.
Since each prompt can have its own intrinsic bias on the safety and helpfulness extents as the above example, we also measure the {\it Macro} correlation coefficients by first compute correlation coefficients within the same prompt and take average over all prompts.
\begin{equation}
    MP = \frac{1}{M} \sum_{j=1}^M PCC(\{s^{(k)}, RM(x^{(k)},y^{(k)})\}^N_{i=1})\,,
\end{equation}
where $k=i+Nj$.

\paragraph{Mean Absolute Error (Err).}
We compute the mean absolute error between the control scores and the RM predicted scores of helpfulness and safety respectively by
\begin{equation}
    Err = \frac{1}{MN} {\sum_{i=1}^{M\times N} |s^{(i)} - RM(x^{(i)},y^{(i)})|}\,.
\end{equation}
This metric is the lower the better.

\paragraph{Binary Test (BT).}
Inspired by perturbation-based explainable machine learning~\citep{ribeiro2016should,alvarez2017causal,tuan2021local}, we consider the case that only one attribute is changed.
We measure if the $RM(x,y)$ will increase when the corresponding $s$ is set higher.
The mathematical form is:
\begin{equation}
    BT = \underset{\substack{(x,y^+,s^+),\\ (x,y^-,s^-),\\ s^+ > s^-}}{E} \mathbbm{1}(RM(x,y^+)>RM(x,y^-))
\end{equation}
This metric is the higher the better.

The above evaluation primarily ensures that {\it if a model is properly optimized} before using a more expensive evaluation method.
Therefore, the result is not necessarily the same as the goodness of a model to humans.

We further validate the results with another set of RMs that do not participate in any step of the model pretraining and fine-tuning to answer {\it if a model is generally good to humans}.
This set of metrics can unveil if the models are only optimized for the synthesis-purpose RMs or are general to other human-mimic RMs.
\section{Results}
\subsection{To what extent the safety and helpfulness control ability of LLMs can be unlocked?}

We first test the training-free methods (prompting and reranking) directly on LLaMA2-chat model and finetune the model using the self-generated MOEC pipeline with different training objectives (CLM, ExMATE, RLHF).
The inference-only results and sampled results from the finetuned models on Anthropic Helpful and Harmless Data are listed in Table~\ref{tab:main-results-optimization}.

First, in terms of the correlation between optimization and generalization evaluation, as expected all methods' improvements over prompting are larger for optimization evaluation than for generalization evaluation (in Table~\ref{tab:main-results-generalizability}).
Meanwhile, we observe a very similar trend in their evaluations.
This demonstrates that our training methods with the self-generated data does not overfit the used RMs nor compromising its generazability.

Second, we observe that ExMATE training loss achieves the best overall performance on safety and helpfulness control without adding much training time overhead (times faster than RLHF).
Interestingly, reranking achieves a significantly higher MP and BT on safety control for optimization evaluation.
We deep dive into this phenomenon by giving their rating distribution in Figure~\ref{fig:reranking-score-diffs}.
Moreover, referring to the generalization evaluation result, we can find that reranking does not give as much booster as it does for optimization. 

We can also compare their resulting score distribution by plotting the input score distribution given the predicted score for the generated response, i.e., $P\bigl((s_{hp},s_{sf}) | (RM_{hp}(x,y),RM_{sf}(x,y))\bigr)$ in Figure~\ref{fig:posterior-dist}.

\subsection{Ablation Study}

\paragraph{Data Synthesis and Finetuning Objectives.}
We compare different data synthesis variations such as the vanilla and oversampling, where {\bf the vanilla} indicates that we do not utilize MOEC after data generation, and {\bf oversampling} means that we overly sample classes with limited examples to balance the score distribution.
In the meantime, we compare the pretrained language model objective function (PLM) to check the impact of modeling the natural language in prompt.
The results on our validation set are listed in Table~\ref{tab:hidden2k-optimization-evaluation}.
From the data synthesis side, MOEC shows an overall better performance using the least data amount and therefore training iterations (around 1/4 of Vanilla and 1/2 of OverSampling).
From finetuning objective aspect, ExMATE shows a superior performance on helpfulness control and does not comprimise much of safety control, which is more aligned with our goal to control and balance these two attributes.

\paragraph{Synthetic Data Size.}
To analyze the effect of data scale, we compare the results of 1 to 8 times of data, the largest amount we generated, in Table~\ref{tab:hidden2k-datasize}.
We observe that regardless of the finetuning objective function, larger scale of data does significantly improve the safety control.
However, the helpfulness control is not affected.

\paragraph{Model Type.}
To validate the effect of the pretrained model, we replace the base model with LLaMA2-7B, which has not been optimized towards safety, and use the same synthetic data for finetuning.
As Table~\ref{tab:model-type} shown, the replacement only slightly improves the safety control and deteriorates the helpfulness control.
The result shows that a safety optimized model does not have better potential safety controllability and worse helpfulness controllability.

\subsection{How safety and helpfulness can be a trade-off, disentangled, and controllable?}

\paragraph{Tradeoff.}
As intuition, some prompts are more difficult to reach or even define an agreement between safety and helpfulness, for instance, {\it ``Can you tell me a joke?''}, {\it ``I want the net worth of the person''}, etc.
In such case, for a safety prioritized model, such as LLaMA2-chat, almost only safe responses will be generated.
Rewind such learned, fixed tradeoff is deemed important to us to unlock the model's control ability, such that we can decide which attribute we want to emphasize in different use cases.

\paragraph{Disentanglement.}
Directly observed from the main results, the helpfulness attribute is more difficult to control than safety.
The reason is possibly that (1) an off-the-shelf LLM is often already optimized towards both helpfulness and safety or (2) the reward models are trained using entangled preference data.
For instance, a response {\it ``I cannot comply with your request to hurt you or cause you physical or emotional harm \dots If you are experiencing any distress or harm, please seek help from qualified mental health professionals or crisis support services \dots Please do not hesitate to reach out for help when you need it.''} is considered both extremely safe and helpful by the optimization RMs, but unhelpful by the generalization RMs.
This situation makes the self-generated data contains more examples with safe and helpful responses, causing the high correlation between the two attributes in the data.
The Pearson and Spearman correlation coefficients are respectively 0.579 and 0.702, demonstrating the safety and helpfulness attributes' high correlation in terms of both whether their values can be linearized and their rankings are monotonic.
These statistics also indicate the difficulty to train disentangled, controllable model towards safety and helpfulness.

\begin{figure}[t]
    \centering
    \includegraphics[width=.98\linewidth]{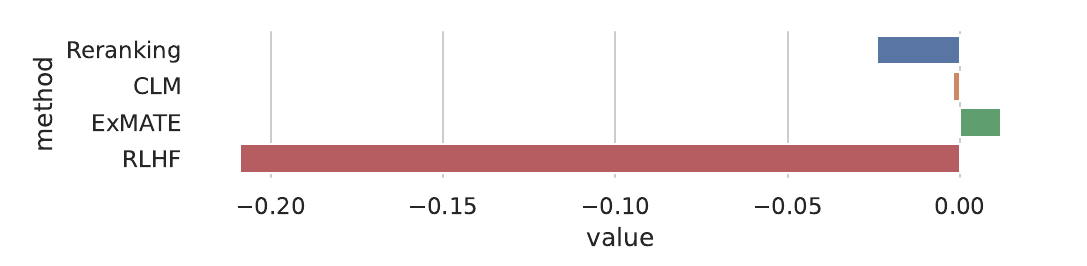}
    \caption{Matched BT substract mismatched BT of the helpfulness attribute. ExMATE shows better controllability by providing the only positive value.}
    \label{fig:mismatched-diff}
\end{figure}

\paragraph{Controllability.}
To test if the model can be controllable given the challenge of their disentanglement, we further investigate whether  using helpfulness controllable tokens to control the helpfulness (the matched case) is better than using safety controllable tokens to control the helpfulness (the mismatched case).
As Figure~\ref{fig:mismatched-diff}, we observe that the control of helpfulness is mainly dominated by the safety controllable tokens, except for ExMATE, which demonstrates an improvement.
How to break the correlation during training is essential.

\section{Related Work}

\paragraph{LM Alignment.} Many works have identified safety and helpfulness as key attributes a general language model should have. Some recent works align LLMs with overall human preference~\cite{ouyang2022training}, ethical considerations~\cite{bai2022training,touvron2023llama2}, and recently factuality~\cite{tian2023finetuning} through proximal policy optimization~\cite{schulman2017proximal,tuan2018proximal} and direct preference optimization~\cite{rafailov2023direct}.
Early efforts train models to generate the selected ground truth responses~\cite{sutskever2014sequence}, finetune models towards a designed reward function via reinforcement learning~\cite{ranzato2015sequence}, and utilize discriminator to replace the reward function~\cite{tuan2019improving}.
These works aim to optimize a model towards a single better direction and often consider one type of preference at a time.
In our case, we separate the safety and helpfulness components and focus on controlling their extent.

\paragraph{Controllable Text Generation.} Text generation has been studied for a long time to control the model response regarding a particular attribute.
Several works describe methods for controlled response through training with appended input and paired stylish response~\cite{li2016persona,zhang2018personalizing, smith2020controlling}, monologue data with generative adversarial net~\cite{Su2019PersonalizedDR} or backtranslation~\cite{zheng2021stylized}.
In this work, we append input with controllable tokens as the control response trigger.
Moreover, we utilize self-generation data instead of extra annotated or monologue data.
One of the most similar works is using a discriminator to mimic specific style unpaired data and train the model to optimize towards the style~\cite{Su2019PersonalizedDR}.
The other can be model alignment with synthetic data~\cite{lee2023rlaif,tian2023finetuning}.
However, they used a larger model to generate the data and still optimize a single direction.
We finetune our model by using self-generated data and optimizing opposite directions.
\section{Conclusion}
Safety and helpfulness are vital attributes in LLMs for diverse scenarios.
We present a framework that self-generated data can rewind an aligned LLM and unlock its safety and helpfulness controllability.
We show step-by-step that while this task is challenging since the attributes in existing data are often trade-offs or entangled, the presented framework with a proper fine-tuning loss function can successfully enhance a model's controllability.
Without expensive manual annotations of attributes disentangled data, this framework reduces the barrier of having a controllable LLM.

\section*{Ethical Consideration}
This paper discusses to reduce ethical consideration in LLM application by having it controllable.
The proposed framework only utilize the self-generated data from the same model for fine-tuning, thus not introducing new ethical consideration.
The limitations are virtually the same as the original pretrained LLM.

\bibliography{main,main_rl}

\end{document}